\documentclass[conference]{IEEEtran}
\IEEEoverridecommandlockouts
\usepackage{cite}
\usepackage{amsmath,amssymb,amsfonts}
\usepackage{algorithmic}
\usepackage{graphicx}
\usepackage{textcomp}
\usepackage{xcolor}
\def\BibTeX{{\rm B\kern-.05em{\sc i\kern-.025em b}\kern-.08em
    T\kern-.1667em\lower.7ex\hbox{E}\kern-.125emX}}

\usepackage{graphicx}
\graphicspath{{./Figures/}}
\usepackage{dcolumn}
\usepackage{bm}
\usepackage{hyperref}

\begin{document}

\title{QUBO Formulations for Training Machine Learning Models\\
\thanks{This manuscript has been authored in part by UT-Battelle, LLC under Contract No. DE-AC05-00OR22725 with the U.S. Department of Energy. The United States Government retains and the publisher, by accepting the article for publication, acknowledges that the United States Government retains a non-exclusive, paid-up, irrevocable, world-wide license to publish or reproduce the published form of this manuscript, or allow others to do so, for United States Government purposes. The Department of Energy will provide public access to these results of federally sponsored research in accordance with the DOE Public Access Plan (http://energy.gov/downloads/doe-public-access-plan). This research used resources of the Oak Ridge Leadership Computing Facility, which is a DOE Office of Science User Facility supported under Contract DE-AC05-00OR22725.}
}

\author{\IEEEauthorblockN{Prasanna Date}
\IEEEauthorblockA{
\textit{Oak Ridge National Laboratory}\\
Tennessee, USA \\
datepa@ornl.gov}    
\and
\IEEEauthorblockN{Davis Arthur}
\IEEEauthorblockA{
\textit{Auburn University}\\
Alabama, USA \\
}
\and
\IEEEauthorblockN{Lauren Pusey-Nazzaro}
\IEEEauthorblockA{
\textit{Washington University in St. Louis}\\
Missouri, USA \\
}
}

\maketitle

\begin{abstract}
Training machine learning models on classical computers is usually a time and compute intensive process.
With Moore's law coming to an end and ever increasing demand for large-scale data analysis using machine learning, we must leverage non-conventional computing paradigms like quantum computing to train machine learning models efficiently.
Adiabatic quantum computers like the D-Wave 2000Q can approximately solve NP-hard optimization problems, such as the quadratic unconstrained binary optimization (QUBO), faster than classical computers.
Since many machine learning problems are also NP-hard, we believe adiabatic quantum computers might be instrumental in training machine learning models efficiently in the post Moore's law era.
In order to solve a problem on adiabatic quantum computers, it must be formulated as a QUBO problem, which is a challenging task in itself.
In this paper, we formulate the training problems of three machine learning models---linear regression, support vector machine (SVM) and equal-sized k-means clustering---as QUBO problems so that they can be trained on adiabatic quantum computers efficiently.
We also analyze the time and space complexities of our formulations and compare them to the state-of-the-art classical algorithms for training these machine learning models.
We show that the time  and space complexities of our formulations are better (in the case of SVM and equal-sized k-means clustering) or equivalent (in case of linear regression) to their classical counterparts.
\end{abstract}


\section{Introduction}
\label{sec:intro}
The importance of machine learning algorithms in scientific advancement cannot be understated.
Machine learning algorithms have given us great predictive power in medical science \cite{obermeyer2016predicting}, economics \cite{yatchew1998nonparametric}, agriculture \cite{mcqueen1995applying} etc. These algorithms can only be implemented and deployed after they have been trained---a process that requires tuning the model parameters of a given machine learning model in order to extract meaningful information from large amounts of data. Training a machine learning model is a time and compute intensive process usually. 
In such situations, one is often forced to make a trade-off between the accuracy of a trained model and the training time.
With the looming end of Moore's law and rapidly increasing demand for large-scale data analysis using machine learning, there is a dire need to explore the applicability of non-conventional computing paradigms like quantum computing to accelerate the training of machine learning models.

Quantum computers are known to bypass classically-difficult computations with great ease by performing operations on high-dimensional tensor product spaces.
To this extent, we believe that machine learning problems, which often require such manipulation of high-dimensional data sets, can be posed in a manner conducive to efficient quantum computation.
Quantum computers have been shown to yield approximate solutions to NP-complete problems, such as the quadratic unconstrained binary optimization (QUBO) problem \cite{date2019efficiently}, graph clustering problem \cite{schaeffer2007graph}, protein folding problem \cite{proteinfolding} etc.
Demonstration of quantum supremacy by Google \cite{arute2019quantum} has led us to believe that quantum computers might offer considerable speedup in a much wider range of use cases such as accelerating training of machine learning models.




In this paper, we formulate the training problems of three machine learning models---linear regression, support vector machine and equal-sized $k$-means clustering---as QUBO problems so that they can be trained on adiabatic quantum computers like D-Wave 2000Q.
The principal contributions of our work are:
\begin{enumerate}
\item We show that the task of training the following machine learning models can be equivalently formulated as a QUBO problem, and thus, efficiently solved using adiabatic quantum computers: linear regression, support vector machine (SVM), and equal-sized $k$-means clustering. 

\item For the aforementioned models, we provide a theoretical comparison between state-of-the-art classical training algorithms and our formulations that are conducive to being trained on adiabatic quantum computers. We observe that the time and space complexities of our formulations are better (in the case of SVM and equal-sized $k$-means clustering) or equivalent (in case of linear regression) to their classical counterparts.

\end{enumerate}


Our formulations provide a promising outlook for training such machine learning models on adiabatic quantum computers. In the future, larger and more robust quantum computers are sought to abate the limitations of current machines and potentially allow machine learning models to be trained faster and more reliably. 

\section{Adiabatic Quantum Computers}
\label{sec:aqc}
The adiabatic theorem states that a quantum physical system remains in its instantaneous eigenstate under a slowly acting perturbation if there is a gap between its eigenvalue and the rest of the Hamiltonian's spectrum \cite{born1928beweis}.
Adiabatic quantum computers leverage the adiabatic theorem to perform computation \cite{farhi2000quantum}.
Specifically, they leverage quantum fluctuations in quantum annealing to find the global minimum of a given objective function over a set of feasible solutions \cite{kadowaki1998quantum}.
The D-Wave adiabatic quantum computers, for instance, are adept at approximately solving the quadratic unconstrained binary optimization (QUBO) problem, which is stated as follows:
\begin{align}
    \min_{z \in \mathbb{B}^M} z^T A z + z^T b \label{eq:qubo}
\end{align}
where, 
$\mathbb{B} = \{0, 1\}$ is the set of binary numbers;
$z \in \mathbb{B}^M$ is the binary decision vector;
$A \in \mathbb{R}^{M \times M}$ is the real-valued $M \times M$ QUBO matrix; and,
$b \in \mathbb{R}^M$ is the real-valued $M$-dimensional QUBO vector.


\section{Notation}
\label{sec:notation}
We use the following notation throughout this paper:
\begin{itemize}
    \item $\mathbb{R}$: Set of real numbers
    \item $\mathbb{B}$: Set of binary numbers, i.e. $\mathbb{B} = \{0, 1\}$.
    \item $\mathbb{N}$: Set of natural numbers
    \item $N$: Number of datapoints (number of rows) in the training dataset
    \item $d$: Number of features (number of columns) in the training dataset
    \item $X$: Training dataset, usually $X \in \mathbb{R}^{N \times (d)}$, i.e. $X$ contains $N$ data points ($N \in \mathbb{N}$) along its rows, and each data point is a $d$ dimensional row vector ($d \in \mathbb{N}$).
    \item $Y$: Classification labels in case of classification tasks.
\end{itemize}

\section{Linear Regression}
\label{sec:regression}

\subsection{Background}
\label{sub:regression-background}

Linear regression is one of the oldest statistical machine learning techniques that is used in a wide range of applications, such as scientific research \cite{leatherbarrow1990using}, business \cite{dielman2001applied}, and weather forecasting \cite{paras2016simple}.
Linear regression models the relationship between a dependent variable and one or more independant variables.

Adiabatic quantum computing approaches have been proposed in the literature for solving the linear regression problem (Equation \ref{eq:regression}).
Borle et al. propose a quantum annealing approach for the linear least squares problem \cite{borle2019analyzing}.
Chang et al. present a quantum annealing approach for solving polynomial systems of equations using least squares \cite{chang2019least}.
Chang et al. propose a method for solving polynomial equations using quantum annealing and discuss its application to linear regression \cite{chang2019quantum}.
These approaches can only find positive real-valued regression weights, while our formulation finds both positive and negative real-valued regression weights.

\begin{figure}
    \centering
    \includegraphics[scale=0.5]{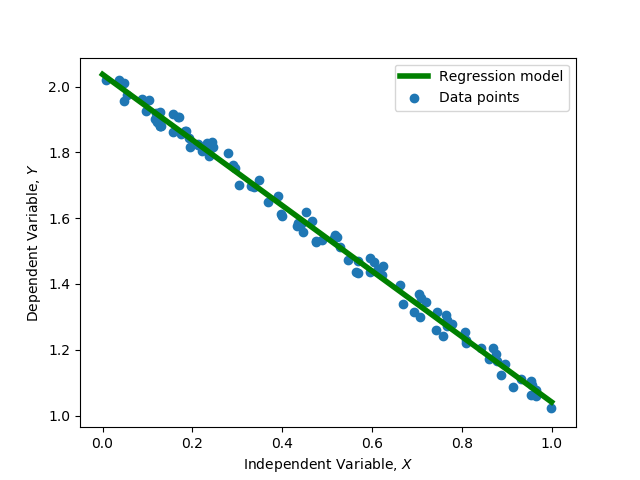}
    \caption{Fitting a linear regression model (green line) to data (blue dots).}
    \label{fig:regression-explained}
\end{figure}

Here, we denote $X \in \mathbb{R}^{N \times (d+1)}$ as the augmented regression training data matrix, where we have augmented each row of the original $X \in \mathbb{R}^{N \times d}$ with unity for the sake of mathematical convenience. 
The regression training labels are denoted by $Y \in \mathbb{R}^{N}$, and the regression weights are denoted by $w \in \mathbb{R}^{d+1}$.
Given $X$ and $Y$, training a linear regression model can be stated as follows:
\begin{align}
    \min_{w \in \mathbb{R}^d} \ E(w) &= || Xw - Y ||^2. \label{eq:regression}
\end{align}
Here, $E(w)$ is the Euclidean error function.
With reference to Figure \ref{fig:regression-explained}, the blue dots represent the data points $X$ and $Y$, and the green line, characterized by the weights $w$, is the regression hyperplane which fits the data.
The regression problem has an analytical solution, given by
\begin{align}
    w = (X^T X)^{-1} X^T Y.
\end{align}
If $(X^T X)^{-1}$ does not exist, the pseudo inverse is computed.
The time complexity of linear regression is known to be $\mathcal{O}(N d^2)$.

\subsection{QUBO Formulation}
\label{sub:regression-formulation}
We start by rewriting Problem \ref{eq:regression} as:
\begin{align}
    \min_{w \in \mathbb{R}^{d+1}} E(w) 
    &= w^T X^T X w - 2 w^T X^T Y + Y^T Y. \label{eq:regression-dotproduct}
\end{align}

Next, we introduce the notion of a $K$-dimensional precision vector $P = [p_1, p_2, \ldots, p_K]^T$. Each entry in $P$ can be an integral power of $2$, and can be both positive or negative. We also introduce a $K$-dimensional vector $\hat{w_i} \in \mathbb{B}^K$ with binary coefficients, such that the inner product $\hat{w_i}^T P$ yields a scalar $w_i \in \mathbb{R}$. This scalar $w_i$ represents the $i^{\text{th}}$ entry in our weight vector, where $1 \le i \le (d+1)$. Note that the entries of $P$ must be sorted, for instance $P = \left[ -2, -1, -\frac{1}{2}, \frac{1}{2}, 1, 2, \right]^T$. 
$\hat{w}_{ik}$ can be thought of as a binary decision variable that selects or ignores entries in $P$ depending on whether its value is $1$ or $0$ respectively.
With this formulation, we can have up to $2^K$ unique values for each $w_i$ when $P$ contains only positive values for instance.
However, if $P$ contains negative values as well, then the number of unique attainable values for each $w_{i}$ might be less than $2^K$.
For example, if $P = [-1, -\frac{1}{2}, \frac{1}{2}, 1]$, then only the following seven distinct values can be attained: $\{-\frac{3}{2}, -1, -\frac{1}{2}, 0, \frac{1}{2}, 1, \frac{3}{2}\}$.

Now, let us define the $K(d+1)$ dimensional binary vector $\hat{w} \in \mathbb{B}^{K(d+1)}$, such that 
\begin{align}
    \hat{w} &= [\hat{w}_{11}, \ldots, \hat{w}_{1K}, \hat{w}_{21}, \ldots, \hat{w}_{2K}, \ldots, \hat{w}_{(d+1)1}, \ldots,  \nonumber \\ 
    & \qquad \hat{w}_{(d+1)K}]^T.
    \label{eq:w-vector}
\end{align}
Similarly, we can define a precision matrix ($\mathcal{P}$) as follows:
\begin{align}
    \mathcal{P} = I_{d+1} \otimes P^T,
    \label{eq:prec-matrix}
\end{align}
where $I_{d+1}$ represents the $(d+1)$-dimensional identity matrix, and $\otimes$ represents the Kronecker product. Note that $\mathcal{P}$ has dimension $(d+1) \times K(d+1)$. We can now recover our original weight vector by observing that: 
\begin{align}
    w &= \mathcal{P} \hat{w}. \label{eq:w-eq-P-w-hat}
\end{align}
We have thus represented our weight vector (to finite precision) in terms of the precision matrix $\mathcal{P}$ and the binary vector $\hat{w} \in \mathbb{B}^{K(d+1)}$. We are now able to pose the minimization problem of Equation \ref{eq:regression-dotproduct} as an equivalent QUBO problem. Let us substitute the expression we obtained for  the weight vector $w$ in terms of $\mathcal{P}$ and $\hat w$ into equation \ref{eq:regression-dotproduct}, which yields:

\begin{align}
    \min_{\hat{w} \in \mathbb{B}^{(d+1)K}} E(\hat{w}) &= \hat{w}^T \mathcal{P}^T X^T X \mathcal{P} \hat{w} - 2 \hat{w}^T \mathcal{P}^T X^T Y. \label{eq:regression-to-qubo}
\end{align}
Note that we have neglected the term $Y^TY$ because it is a constant scalar and does not affect the solution to this unconstrained optimization problem. Observe that Equation \ref{eq:regression-to-qubo} now has the form of a QUBO problem, as desired. Hence, we are able to solve this optimization problem using an adiabatic quantum computer.

\subsection{Theoretical Analysis}
\label{sub:regression-analysis}
The regression problem (Problem \ref{eq:regression}) has $\mathcal{O}(N d)$ data ($X$ and $Y$) and $\mathcal{O}(d)$ weights ($w$), which is the same for Problem \ref{eq:regression-to-qubo}.
We introduced $K$ binary variables for each of the $d+1$ weights when converting Problem \ref{eq:regression} to Problem \ref{eq:regression-to-qubo}.
So, we have $\mathcal{O}(d K)$ variables in Equation \ref{eq:regression-to-qubo}, which translates to quadratic qubit footprint ($\mathcal{O}(K^2 d^2)$) using an efficient embedding algorithm such as \cite{date2019efficiently}. 
Embedding is the process of mapping logical QUBO variables to qubits on the hardware, and is challenging because inter-qubit connectivity on the hardware is extremely limited.
As mentioned in Section \ref{sub:regression-background}, solving the regression problem (Equation \ref{eq:regression}) takes $\mathcal{O}(N d^2)$ time classically.
From Equation \ref{eq:regression-to-qubo}, we can infer that the QUBO formulation takes $\mathcal{O}(N d^2 K^2)$ time.
Obtaining the solution on adiabatic quantum computers depends on the annealing time, which is not $\mathcal{O}(1)$ in general, but can be treated as $\mathcal{O}(1)$ for all practical purposes \cite{date2019classical}.
So, the total time to convert and solve a linear regression problem on adiabatic quantum computer would be $\mathcal{O}(N d^2 K^2)$.

It is clear that this running time is worse than its classical counterpart ($\mathcal{O}(N d^2)$).
However, the above analysis assumes that $K$ is variable.
On classical computers, the precision is fixed, for example, $32$-bit or $64$-bit precision.
We can analogously fix the precision for quantum computers, and take $K$ to be a constant.
The resulting qubit footprint would be $\mathcal{O}(d^2)$, and the time complexity would be $\mathcal{O}(N d^2)$, which is equivalent to the classical algorithm.

\section{Support Vector Machine (SVM)}
\label{sec:svm}

\subsection{Background}
\label{sub:svm-background}
\begin{figure}
    \centering
    \includegraphics[scale=0.5]{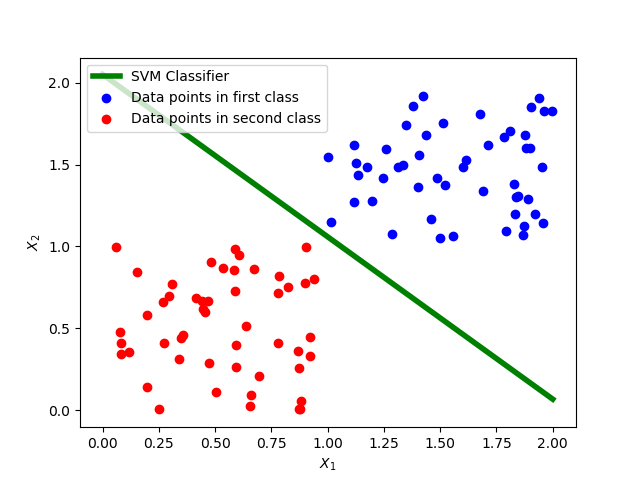}
    \caption{SVM model (green line) correctly classifying training data (red and blue dots).}
    \label{fig:svm-explained}
\end{figure}

Support vector machine (SVM) is a powerful supervised machine learning model that produces robust classifiers as shown in Figure \ref{fig:svm-explained}.
The classifier produced by SVM maximizes its distance from the classes of the data points.
Although SVM was meant for binary classification originally, several variants of SVM have been proposed over the years that allow multi-class classification \cite{bo2006svm,cheong2004support}.
SVM has wide ranging applications in multimedia (vision, text, speech etc.) \cite{moreno2004kullback}, biology \cite{byvatov2003support}, and chemistry \cite{ivanciuc2007applications}, among many scientific disciplines.

Quantum approaches for training SVM using adiabatic quantum computers have been proposed in the literature.
Ahmed proposes a formulation for quantum SVM that runs on noisy intermediate-scale quantum (NISQ) processors \cite{ahmed2019pattern}.
Welsh et al. propose a formulation of SVM for the D-Wave quantum computers \cite{willsch2020support}.
Our findings improve upon their formulation, allowing for real-valued learning parameters up to a certain precision.

Given training data $X \in \mathbb{R}^{N \times d}$ and training labels $Y \in \{-1, +1\}^N$, we would like to find a classifier (determined by weights, $w \in \mathbb{R}^d$, and bias, $b \in \mathbb{R}$), that separates the training data. 
Formally, training SVM is expressed as:
\begin{align}
    & \min_{w, b} \ ||w||^2 \label{eq:svm-qpp} \\
    \text{subject to:} \quad & y_i (w^T x_i + b) \ge 1 \qquad \forall i = 1, 2, \ldots, N \nonumber.
\end{align}

Note that the objective function is convex because its Hessian matrix is the identity matrix, which is positive definite.
Furthermore, since the constraints are linear, they are convex as well, which makes Problem \ref{eq:svm-qpp} a convex quadratic programming problem.
To solve Problem \ref{eq:svm-qpp}, we first compute the Lagrangian dual as follows:
\begin{align}
    \max_{w, b, \lambda} \ \mathcal{L}(w, b, \lambda) = ||w||^2 - \sum_{i = 1}^{N} \lambda_i \left[ y_i (w^T x_i + b) - 1 \right], \label{eq:lagrangian-dual}
\end{align}

where $\lambda$ is the vector containing all the Lagrangian multipliers, i.e. $\lambda = [\lambda_1 \ \lambda_2 \ \cdots \ \lambda_N]^T$, with $\lambda_i \ge 0 \ \forall i$. 
The non-zero Lagrangian multipliers in the final solution correspond to the support vectors and determine the hyperplanes $H_1$ and $H_2$ in Figure \ref{fig:svm-explained}.
The Lagrangian dual problem (Equation \ref{eq:lagrangian-dual}) is solved in $\mathcal{O}(N^3)$ time on classical computers by applying the Karush-Kuhn-Tucker (KKT) conditions \cite{karush1939minima,kuhn2014nonlinear}.

\subsection{QUBO Formulation}
\label{sub:svm-formulation}
In order to convert SVM training into a QUBO problem, we write Equation \ref{eq:lagrangian-dual} as a minimization problem:
\begin{align}
    \min_{w, b, \lambda} \mathcal{L}(w, b, \lambda) &= - w^T w + w^T (X \odot Y')^T \lambda + b Y^T \lambda - 1_{N}^T \lambda \label{eq:lagrangian-dual-min}
\end{align}

where $Y'$ represents the $N \times d$ matrix obtained by stacking $Y$ horizontally $d$ times, i.e. $Y' = [Y \ Y \ \cdots \ d\text{ times}]$;
$\odot$ is the element-wise multiplication operation;
and, $1_N$ represents an $N$-dimensional vector of ones.
Next, we define the variable vector $\theta$, matrix $U$ and vector $v$ as follows:
\begin{align}
    \theta &=   \begin{bmatrix}
                    w \\ b \\ \lambda
                \end{bmatrix}, \
    U =    \begin{bmatrix}
                -I_d & 0 & (X \odot Y')^T \\
                0   & 0 & Y^T \\
                0   & 0 & 0
            \end{bmatrix}, \
    v =    -\begin{bmatrix}
                0 \\ 0 \\ 1_N,
            \end{bmatrix} \label{eq:svm-theta-U-v} 
\end{align}
where $I_d$ is a $d$ dimensional identity matrix.
Now, we can rewrite Equation \ref{eq:lagrangian-dual-min} in matrix form as follows:
\begin{align}
    \min_{\theta} \mathcal{L}(\theta) &= \theta^T U \theta + \theta^T v. \label{eq:lagrangian-dual-matrix}
\end{align}

We now reintroduce the $K$-dimensional precision vector $P = [p_1, p_2, \ldots, p_K]^T$, as described in Section \ref{sub:regression-formulation}.
Next, we introduce $K$ binary variables $\hat{w}_{jk}$, $\hat{b}_{k}$, $\hat{\lambda}_{ik}$ for each SVM weight, bias and Lagrangian multiplier:
\begin{align}
    w_{j} &= \sum_{k=1}^K p_k \hat{w}_{jk} \qquad \forall j = 1, 2, \ldots, d \label{eq:binarized-svm-weights} \\
    b &= \sum_{k=1}^K p_k \hat{b}_{k} \label{eq:binarized-svm-bias} \\
    \lambda_i &= \sum_{k=K_+}^K p_k \hat{\lambda}_{ik} \qquad \forall i = 1, 2, \ldots, N \label{eq:binarized-svm-lagrange-multipliers}
\end{align}

where, $p_k$ denotes the $k^{th}$ entry in the precision vector $P$;
and, $K_+$ denotes the index of smallest positive entry in $P$.
Summing from $K_+$ in Equation \ref{eq:binarized-svm-lagrange-multipliers} ensures that the Lagrange multipliers are always positive, which is required when solving the Lagrangian dual (Problem \ref{eq:lagrangian-dual-min}).

Now, we vertically stack all binary variables as follows:
\begin{align}
    \hat{w} &= [\hat{w}_{11} \ldots \hat{w}_{1K} \ \hat{w}_{21} \ldots \hat{w}_{2K} \ \ldots \ \hat{w}_{d1} \ldots \hat{w}_{dK} ]^T \label{eq:w-hat} \\
    \hat{b} &= [\hat{b}_{1} \ldots \hat{b}_K]^T \label{eq:b-hat} \\
    \hat{\lambda} &= [\hat{\lambda}_{1K_+} \ldots \hat{\lambda}_{1K} \ \hat{\lambda}_{2K_+} \ldots \hat{\lambda}_{2K} \ \ldots \ \hat{\lambda}_{NK_+} \ldots \hat{\lambda}_{NK} ]^T \label{eq:lambda-hat}
\end{align}

We also define the precision matrix as follows:
\begin{align}
    \mathcal{P} &=  \begin{bmatrix}
                        I_{d+1} \otimes P^T     & 0_{(d+1) \times N (K - K_+ + 1)} \\
                        0_{N \times (d + 1)}    & I_N \otimes P_+^T 
                    \end{bmatrix} \label{eq:precision-matrix}
\end{align}

where,
$0_{I \times J}$ denotes $I \times J$ matrix of zeroes;
$P_+$ denotes the vector containing only the positive elements in P.
The dimensions of the resulting $\mathcal{P}$ are $(N + d + 1) \times (K(d + 1) + N (K - K_+ + 1))$.
Equations \ref{eq:w-hat}, \ref{eq:b-hat}, \ref{eq:lambda-hat} and \ref{eq:precision-matrix} are done for mathematical convenience.
Now, we stack $\hat{w}$, $\hat{b}$ and $\hat{\lambda}$ as the vector $\hat{\theta}$:
\begin{align}
    \hat{\theta} =  \begin{bmatrix}
                        \hat{w} \\ \hat{b} \\ \hat{\lambda}
                    \end{bmatrix} \label{eq:theta-hat}
\end{align}
Notice that:
\begin{align}
    \theta &= \mathcal{P} \hat{\theta} \label{eq:theta-theta-hat}
\end{align}

Finally, we substitute the value of $\theta$ from Equation \ref{eq:theta-theta-hat} into Equation \ref{eq:lagrangian-dual-matrix}:
\begin{align}
    \min_{\hat{\theta}} \mathcal{L} (\theta) &= \hat{\theta}^T \mathcal{P}^T U \mathcal{P} \hat{\theta} + \hat{\theta}^T  \mathcal{P}^T v \label{eq:svm-qubo}
\end{align}

Equation \ref{eq:svm-qubo} is identical to Equation \ref{eq:qubo} with $z = \hat{\theta}$, $A = \mathcal{P}^T U \mathcal{P}$, $b = \mathcal{P}^T v$, and $M = K (N + d + 1)$.
Hence, we have converted the SVM training problem from Equation \ref{eq:lagrangian-dual} into a QUBO problem in Equation \ref{eq:svm-qubo}, which can be solved on adiabatic quantum computers.

\subsection{Theoretical Analysis}
\label{sub:svm-analysis}
We begin our theoretical analysis by defining the space complexity with respect to the number of qubits needed to solve the QUBO. 
The SVM training problem stated in Equation \ref{eq:lagrangian-dual} contains $\mathcal{O}(N + d)$ variables ($w$, $b$ and $\lambda$) and $\mathcal{O}(Nd)$ data ($X$ and $Y$).
The QUBO formulation of the SVM training problem stated in Equation \ref{eq:svm-qubo} consists of the same amount of data.
However, as part of the QUBO formulation, we introduced $K$ binary variables for each variable in the original problem (Equation \ref{eq:lagrangian-dual}).
So, the total number of variables in Equation \ref{eq:svm-qubo} is $\mathcal{O}(KN + Kd)$.
So, the qubit footprint (or space complexity) of this formulation would be $\mathcal{O}((KN + Kd)^2)$ after embedding onto the hardware.
In a practical setting, the number of data points is larger than the dimension of each data point, i.e. $N \gg d$. 
Thus, the number of variables would be $\mathcal{O}(NK)$, and the qubit footprint would be $\mathcal{O}(N^2 K^2)$.

The time complexity of classical SVM algorithms is known to be $\mathcal{O}(N^3)$ \cite{bottou2007support}.
To compute the time complexity for converting Problem \ref{eq:svm-qpp} into a QUBO problem, we can rewrite Equation \ref{eq:svm-qubo} as follows:
\begin{align}
    \min_{\hat{w}, \hat{b}, \hat{\lambda}} \mathcal{L}(\hat{w}, \hat{b}, \hat{\lambda}) &= -\sum_{j=1}^d \sum_{k=1}^{K} \sum_{l=1}^{K} p_k p_l \hat{w}_{jk} \hat{w}_{jl} \nonumber \\
    & + \sum_{i=1}^{N} \sum_{j=1}^{d} \sum_{k=1}^{K} \sum_{l=K_+}^{K} x_{ij} y_i p_k p_l \hat{w}_{jk} \hat{\lambda}_{il} \nonumber \\
    & + \sum_{i=1}^{N} \sum_{k=1}^{K} \sum_{l=K_+}^{K} y_i p_k p_l \hat{b}_k \hat{\lambda}_{il}
     - \sum_{i=1}^{N} \sum_{l=K_+}^K p_l \hat{\lambda}_{il} \label{eq:svm-qpp-summation}
\end{align}
From Equation \ref{eq:svm-qpp-summation}, the time complexity is $\mathcal{O}(NdK^2)$, which is dominated by the second term.
The process of obtaining the actual solution on the adiabatic quantum computer through quantum annealing can be treated as a constant ($\mathcal{O}(1)$) for all practical purposes.
So, the total time complexity is $\mathcal{O}(N d K^2)$.

Note that the qubit footprint $\mathcal{O}(N^2K^2)$ and time complexity $\mathcal{O}(NdK^2)$ assume that $K$, which is the length of the precision vector is a variable.
If the precision for all parameters ($\hat{w},\hat{b},\hat{\lambda}$) is fixed (e.g. limited to $32$-bit or $64$-bit precision), then $K$ becomes a constant factor.
The resulting qubit footprint would be $\mathcal{O}(N^2)$, and time complexity would be $\mathcal{O}(N d)$.
This time complexity is better than the classical algorithm ($\mathcal{O}(N^3)$).

\section{Equal Size $k$-Means Clustering}
\label{sec:kmeans}

\subsection{Background}
\label{sub:kmeans-background}

\begin{figure}
    \centering
    \includegraphics[scale=0.5]{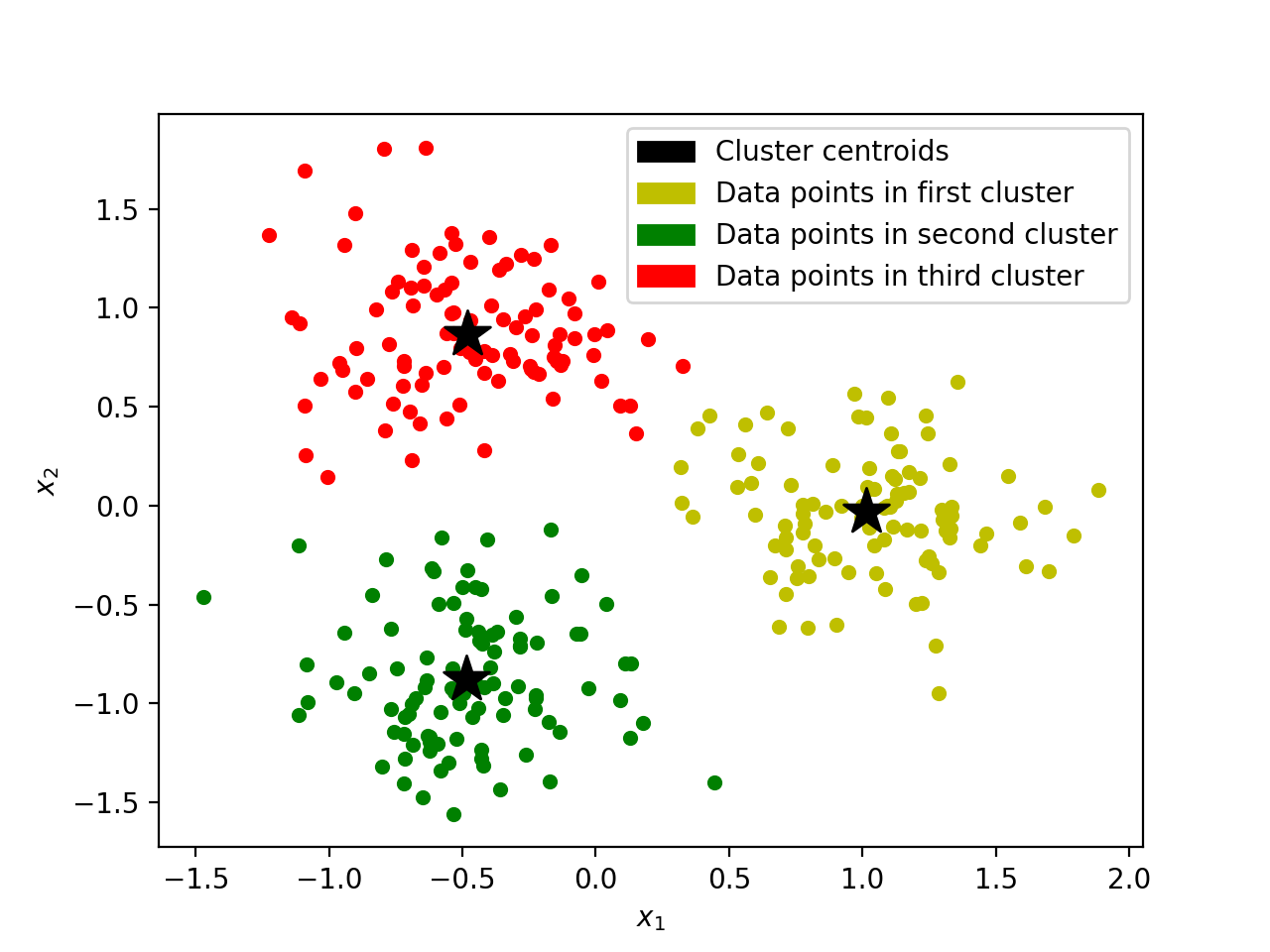}
    \caption{Training an equal size k-means clustering model ($k = 3$) on training data (yellow, green, and red dots).}
    \label{fig:kmeans-explained}
\end{figure}

$k$-Means clustering is an unsupervised learning model that partitions training data into $k$ clusters such that each point belongs to the cluster with the nearest centroid. The optimal cluster assignments of the training data minimizes within cluster variance. Equal size $k$-means clustering is a special case of the $k$-means model with the additional constraint that each cluster contains approximately $N/k$ points as shown in Figure \ref{fig:kmeans-explained}. 
Balanced clustering models have applications in a variety of domains including network design \cite{gupta2003networks}, marketing \cite{Ghosh2005marketing}, and document clustering \cite{Banerjee2003documentclustering}.

Quantum approaches to training clustering models have been discussed in the literature. Ushijima-Mwesigwa et al. demonstrate partitioning a graph into $k$ parts concurrently using quantum annealing on the D-Wave 2X machine \cite{UshijimaMwesigwa2017partitioning}. Kumar et al. present a QUBO formulation for $k$-clustering that differs from the $k$-means model \cite{Kumar2018combinatorialclustering}. Bauckhage et al. propose a QUBO formulation for binary clustering ($k = 2$) \cite{bauckhage2018adiabatic} and $k$-medoids clustering \cite{bauckhage2019qubo}. Our QUBO formulation for equal size $k$-means clustering synthesizes a number of ideas proposed in the literature.

Given training data $X \in \mathbb{R}^{N \times d}$, we would like to partition the $N$ data points into $k$ clusters $\Phi = \{\phi_1, ..., \phi_k\}$. Let the centroid of cluster $\phi_i$ be denoted as $\mu_i$. Formally, training the generic $k$-means clustering model is expressed as:
This minimization problem can also be expressed as:
\begin{align}
    \min_{\Phi} \sum_{i = 1}^k \frac{1}{2|\phi_i|} \sum_{x, y \in \phi_i} || x - y ||^2 \label{eq:kclusters2}
\end{align}
In the case that each cluster is of equal size, $|\phi_i|$ is constant, and Problem \ref{eq:kclusters2} reduces to:
\begin{align}
    \min_{\Phi} \sum_{i = 1}^k \sum_{x, y \in \phi_i} || x - y ||^2
    \label{eq:kmeanssamesize}
\end{align}
Note that for most applications of balanced clustering, the cluster sizes are only approximately equal to one another. In these cases, the solution to Problem \ref{eq:kmeanssamesize} may not be the exact solution to Problem \ref{eq:kclusters2}.

Classically, the $k$-means clustering problem is solved heuristically through an iterative approach known as Lloyd's algorithm. A modified version of this algorithm is used for equal size $k$-means clustering to uphold the constraint that no cluster contains more than $N / k$ points \cite{Ganganath2014samesizekclustering}. This modified version of Lloyd's algorithm runs in $\mathcal{O}(N^{3.5} k^{3.5} )$ time on classical computers \cite{Malinen2014balancedcomplexity}. 

\subsection{QUBO Formulation}
\label{sub:kmeans-formulation}

To formulate Problem \ref{eq:kmeanssamesize} as a QUBO problem, it will be useful to define a matrix $D \in \mathbb{R}^{N \times N}$ where each element is given by:
\begin{align}
    d_{ij} = ||x_i - x_j||^2
\end{align}
We also define a binary matrix $\hat{W} \in \mathbb{B}^{N \times k}$ such that $\hat{w}_{ij} = 1$ iff point $x_i$ belongs to cluster $\phi_j$. Since we are assuming clusters of the same size, each column in $\hat{W}$ should have approximately $N / k$ entries equal to 1. Additionally, since each data point belongs to exactly one cluster, each row in $\hat{W}$ must contain exactly one entry equal to 1. Using this notation, the inner sum in Problem \ref{eq:kmeanssamesize} can be rewritten:
\begin{align}
    \sum_{x, y \in \phi_j} || x - y ||^2 = \hat{w}{'}_j^T D \hat{w}'_j
\end{align}
where $\hat{w}'_j$ is the $jth$ column in $\hat{W}$. From this relation, we can cast Problem \ref{eq:kmeanssamesize} into a constrained binary optimization problem.
First, we vertically stack the $Nk$ binary variables in $\hat{W}$ as follows:
\begin{align}
    \hat{w} = 
    [\hat{w}_{11} \ldots \hat{w}_{N1} \ \hat{w}_{12} \ldots \hat{w}_{N2} \ldots \hat{w}_{1k} \ldots \hat{w}_{Nk}]^T \label{eq:kmeansz}
\end{align}
Provided the constraints on $\hat{w}$ are upheld, Problem \ref{eq:kmeanssamesize} is equivalent to:
\begin{align}
    \min_{\hat{w}} \hat{w}^T (I_k \otimes D) \hat{w}
    \label{eq:kmeansconstrained}
\end{align}
where $I_k$ is the $k$-dimensional identity matrix.

We can remove the constraints on $\hat{w}$ by including penalty terms that are minimized when all conditions are satisfied. First, we account for the constraint that each cluster must contain approximately $N/k$ points. For a given column $\hat{w}'_j$ in $\hat{W}$, this can be enforced by including a penalty of the form:
\begin{align}
    \alpha (\hat{w}{'}_j^T \hat{w}'_j - N/k)^2 
    \label{eq:krow1}
\end{align}
where $\alpha$ is a constant factor intended to make the penalty large enough that the constraint is always upheld. Dropping the constant term $\alpha(N/k)^2$, this penalty is equivalent to $\hat{w}{'}_j^T \alpha F \hat{w}'_j$ where $F$ is defined as:
\begin{align}
    F = 1_N - \frac{2N}{k} I_N
\end{align}
Using this formulation, the sum of all column constraint penalties is:
\begin{align}
    \hat{w}^T (I_k \otimes \alpha F ) \hat{w} \label{eq:kmeanscolumns}
\end{align}

Next, we account for the constraint that each point belongs to exactly $1$ cluster. For a given row $\hat{w}_i$, this can be enforced by including a penalty of the form:
\begin{align}
    \beta (\hat{w}_i^T \hat{w}_i - 1)^2
\end{align}
where $\beta$ is a constant with the same purpose as $\alpha$ in Equation \ref{eq:krow1}. Dropping the constant term, this penalty is equivalent to $\hat{w}_i^T \beta G \hat{w}_i$ where $G$ is defined as:
\begin{align}
    G = 1_k - 2 I_k
\end{align} 
To find the sum of all row constraint penalties, we first convert the binary vector $\hat{w}$ into the form $\hat{v}$ shown below:
\begin{align}
    \hat{v} = [w_{11} \ldots w_{1k} \ w_{21} \ldots w_{2k} \ldots w_{N1} \ldots w_{Nk}]^T
\end{align}
This can be accomplished through a linear transformation $Q \hat{w}$ where each element in $Q \in \mathbb{B}^{Nk \times Nk}$ is defined as:
\begin{align}
    q_{ij} = 
    \begin{cases} 
      1 & j =  N \operatorname{mod}(i - 1, k) + \lfloor \frac{i - 1}{k} \rfloor + 1 \\
      0 & \text{else} \\
   \end{cases}
\end{align}
After the transformation, the sum of all row constraint penalties is given by $\hat{v}^T (I_N \otimes \beta G) \hat{v}$. This can be equivalently expressed as:
\begin{align}
    \hat{w}^T Q^T (I_N \otimes \beta G) Q \hat{w} \label{eq:kmeansrow}
\end{align}
Combining the penalties from Equation \ref{eq:kmeanscolumns} and Equation \ref{eq:kmeansrow} with the constrained binary optimization problem from Equation \ref{eq:kmeansconstrained}, Problem \ref{eq:kmeanssamesize} can be rewritten as:
\begin{align}
    \min_{\hat{w}} \hat{w}^T (I_k \otimes (D + \alpha F) + Q^T (I_N \otimes \beta G) Q) \hat{w} \label{eq:finalkmeans}
\end{align}
Equation \ref{eq:finalkmeans} is identical to Equation \ref{eq:qubo} with $z = \hat{w}$, $A = (I_k \otimes (D + \alpha F) + Q^T (I_N \otimes \beta G) Q) $, and $b = 0$. Thus, we have converted Equation \ref{eq:kmeanssamesize} into a QUBO problem which can be solved on adiabatic quantum computers.

\subsection{Theoretical Analysis}
\label{sub:kmenas-analysis}
The equal size $k$-means clustering problem stated in Equation \ref{eq:kmeanssamesize} contains $\mathcal{O}(Nd)$ data and $\mathcal{O}(N)$ variables. In our QUBO formulation, we introduce $k$ binary variables for each variable in the original problem. Thus, the total number of variables in Equation \ref{eq:finalkmeans} is $\mathcal{O}(Nk)$. This translates to a quadratic qubit footprint of $\mathcal{O}(N^2 k^2)$ using an efficient embedding algorithm such as \cite{date2019efficiently}.

While an exact solution to the generic $k$-means clustering model (Problem \ref{eq:kclusters2}) requires $\mathcal{O}(N^{kd + 1})$ time \cite{Inaba1994kworstcase}, a classical algorithm for equal size $k$-means clustering will converge to a locally optimal solution in $\mathcal{O}(N^{3.5} k^{3.5} )$ time \cite{Malinen2014balancedcomplexity}. To compute the time complexity for converting Equation \ref{eq:kmeanssamesize} into a QUBO problem, we can rewrite Equation \ref{eq:finalkmeans} as follows:
\begin{align}
    \min_{W} & \sum_{l = 1}^k \sum_{j = 1}^N \sum_{i = 1}^N  \sum_{m = 1}^d w_{il} (x_{im} - x_{jm})^2 w_{jl} \nonumber \\
    &+ \alpha \sum_{l = 1}^k \sum_{j = 1}^N \sum_{i = 1}^N w_{il} f_{ij} w_{jl}
    + \beta \sum_{l = 1}^N \sum_{j = 1}^k \sum_{i = 1}^k w_{li} g_{ij} w_{lj}
    \label{eq:kmeanscomplexity}
\end{align}
From Equation \ref{eq:kmeanscomplexity}, the time complexity is $\mathcal{O}(N^2 k d)$, which is dominated by the first term. For practical purposes, solving the QUBO problem through quantum annealing can be done in constant time. Therefore, the total time complexity for the quantum algorithm is $\mathcal{O}(N^2 k d)$. This time complexity is better than the worst case time complexity of the classical algorithm $(\mathcal{O}(N^{3.5} k^{3.5}))$. 
However, the number of iterations in the classical algorithm varies greatly depending on the quality of the initial guess at the cluster centroids. In many cases, the classical algorithm will converge in much less than $\mathcal{O}(N^{3.5} k^{3.5})$ time and outperform its quantum counterpart.

\section{Conclusion}
\label{sec:conclusion}
As the task of training machine learning models becomes more computationally intensive, devising new methods for efficient training has become a crucial pursuit in machine learning. 
The process of training a given model can often be formulated as a problem of minimizing a well-defined error function for a given machine learning model.
Given the power of quantum computers to approximately solve certain hard optimization problems with great efficiency as well as the recent demonstration of quantum supremacy, we believe quantum computers can accelerate training of machine learning models.
In this paper, we posed the training problems for three machine learning models (linear regression, support vector machine, and equal-sized $k$-means clustering) as QUBO problems to be solved on adiabatic quantum computers like D-Wave 2000Q. 
Furthermore, we analyzed the associated time and space complexity of our formulations and provided a theoretical comparison to the state-of-the-art classical methods for training these models. 
Our results are promising for training machine learning models on quantum computers in the future. 

In the future, we would like to empirically evaluate the performance of our quantum approaches on real quantum computers.
We would also like to compare the performance of our quantum approaches to state-of-the-art classical approaches.
Finally, we would like to formulate other machine learning models such as logistic regression, restricted Boltzmann machines, deep belief networks, Bayesian learning and deep learning as QUBO problems that could potentially be trained on adiabatic quantum computers.

\bibliographystyle{IEEEtran}
\bibliography{references}

\begin{thebibliography}{10}
\providecommand{\url}[1]{#1}
\csname url@samestyle\endcsname
\providecommand{\newblock}{\relax}
\providecommand{\bibinfo}[2]{#2}
\providecommand{\BIBentrySTDinterwordspacing}{\spaceskip=0pt\relax}
\providecommand{\BIBentryALTinterwordstretchfactor}{4}
\providecommand{\BIBentryALTinterwordspacing}{\spaceskip=\fontdimen2\font plus
\BIBentryALTinterwordstretchfactor\fontdimen3\font minus
  \fontdimen4\font\relax}
\providecommand{\BIBforeignlanguage}[2]{{%
\expandafter\ifx\csname l@#1\endcsname\relax
\typeout{** WARNING: IEEEtran.bst: No hyphenation pattern has been}%
\typeout{** loaded for the language `#1'. Using the pattern for}%
\typeout{** the default language instead.}%
\else
\language=\csname l@#1\endcsname
\fi
#2}}
\providecommand{\BIBdecl}{\relax}
\BIBdecl

\bibitem{obermeyer2016predicting}
Z.~Obermeyer and E.~J. Emanuel, ``Predicting the future—big data, machine
  learning, and clinical medicine,'' \emph{The New England journal of
  medicine}, vol. 375, no.~13, p. 1216, 2016.

\bibitem{yatchew1998nonparametric}
A.~Yatchew, ``Nonparametric regression techniques in economics,'' \emph{Journal
  of Economic Literature}, vol.~36, no.~2, pp. 669--721, 1998.

\bibitem{mcqueen1995applying}
R.~J. McQueen, S.~R. Garner, C.~G. Nevill-Manning, and I.~H. Witten, ``Applying
  machine learning to agricultural data,'' \emph{Computers and electronics in
  agriculture}, vol.~12, no.~4, pp. 275--293, 1995.

\bibitem{date2019efficiently}
P.~Date, R.~Patton, C.~Schuman, and T.~Potok, ``Efficiently embedding qubo
  problems on adiabatic quantum computers,'' \emph{Quantum Information
  Processing}, vol.~18, no.~4, p. 117, 2019.

\bibitem{schaeffer2007graph}
S.~E. Schaeffer, ``Graph clustering,'' \emph{Computer science review}, vol.~1,
  no.~1, pp. 27--64, 2007.

\bibitem{proteinfolding}
\BIBentryALTinterwordspacing
K.~A. Dill, S.~B. Ozkan, M.~S. Shell, and T.~R. Weikl, ``The protein folding
  problem,'' \emph{Annual Review of Biophysics}, vol.~37, no.~1, pp. 289--316,
  2008, pMID: 18573083. [Online]. Available:
  \url{https://doi.org/10.1146/annurev.biophys.37.092707.153558}
\BIBentrySTDinterwordspacing

\bibitem{arute2019quantum}
F.~Arute, K.~Arya, R.~Babbush, D.~Bacon, J.~C. Bardin, R.~Barends, R.~Biswas,
  S.~Boixo, F.~G. Brandao, D.~A. Buell \emph{et~al.}, ``Quantum supremacy using
  a programmable superconducting processor,'' \emph{Nature}, vol. 574, no.
  7779, pp. 505--510, 2019.

\bibitem{born1928beweis}
M.~Born and V.~Fock, ``Beweis des adiabatensatzes,'' \emph{Zeitschrift f{\"u}r
  Physik}, vol.~51, no. 3-4, pp. 165--180, 1928.

\bibitem{farhi2000quantum}
E.~Farhi, J.~Goldstone, S.~Gutmann, and M.~Sipser, ``Quantum computation by
  adiabatic evolution,'' \emph{arXiv preprint quant-ph/0001106}, 2000.

\bibitem{kadowaki1998quantum}
T.~Kadowaki and H.~Nishimori, ``Quantum annealing in the transverse ising
  model,'' \emph{Physical Review E}, vol.~58, no.~5, p. 5355, 1998.

\bibitem{leatherbarrow1990using}
R.~J. Leatherbarrow, ``Using linear and non-linear regression to fit
  biochemical data,'' \emph{Trends in biochemical sciences}, vol.~15, no.~12,
  pp. 455--458, 1990.

\bibitem{dielman2001applied}
T.~E. Dielman, \emph{Applied regression analysis for business and
  economics}.\hskip 1em plus 0.5em minus 0.4em\relax Duxbury/Thomson Learning
  Pacific Grove, CA, 2001.

\bibitem{paras2016simple}
S.~M. Paras \emph{et~al.}, ``A simple weather forecasting model using
  mathematical regression,'' \emph{Indian research journal of extension
  education}, vol.~12, no.~2, pp. 161--168, 2016.

\bibitem{borle2019analyzing}
A.~Borle and S.~J. Lomonaco, ``Analyzing the quantum annealing approach for
  solving linear least squares problems,'' in \emph{International Workshop on
  Algorithms and Computation}.\hskip 1em plus 0.5em minus 0.4em\relax Springer,
  2019, pp. 289--301.

\bibitem{chang2019least}
T.~H. Chang, T.~C. Lux, and S.~S. Tipirneni, ``Least-squares solutions to
  polynomial systems of equations with quantum annealing,'' \emph{Quantum
  Information Processing}, vol.~18, no.~12, p. 374, 2019.

\bibitem{chang2019quantum}
C.~C. Chang, A.~Gambhir, T.~S. Humble, and S.~Sota, ``Quantum annealing for
  systems of polynomial equations,'' \emph{Scientific reports}, vol.~9, no.~1,
  pp. 1--9, 2019.

\bibitem{date2019classical}
P.~Date, C.~Schuman, R.~Patton, and T.~Potok, ``A classical-quantum hybrid
  approach for unsupervised probabilistic machine learning,'' in \emph{Future
  of Information and Communication Conference}.\hskip 1em plus 0.5em minus
  0.4em\relax Springer, 2019, pp. 98--117.

\bibitem{bo2006svm}
G.~Bo and H.~Xianwu, ``Svm multi-class classification,'' \emph{Journal of Data
  Acquisition \& Processing}, vol.~21, no.~3, pp. 334--339, 2006.

\bibitem{cheong2004support}
S.~Cheong, S.~H. Oh, and S.-Y. Lee, ``Support vector machines with binary tree
  architecture for multi-class classification,'' \emph{Neural Information
  Processing-Letters and Reviews}, vol.~2, no.~3, pp. 47--51, 2004.

\bibitem{moreno2004kullback}
P.~J. Moreno, P.~P. Ho, and N.~Vasconcelos, ``A kullback-leibler divergence
  based kernel for svm classification in multimedia applications,'' in
  \emph{Advances in neural information processing systems}, 2004, pp.
  1385--1392.

\bibitem{byvatov2003support}
E.~Byvatov and G.~Schneider, ``Support vector machine applications in
  bioinformatics.'' \emph{Applied bioinformatics}, vol.~2, no.~2, pp. 67--77,
  2003.

\bibitem{ivanciuc2007applications}
O.~Ivanciuc \emph{et~al.}, ``Applications of support vector machines in
  chemistry,'' \emph{Reviews in computational chemistry}, vol.~23, p. 291,
  2007.

\bibitem{ahmed2019pattern}
S.~Ahmed, ``Pattern recognition with quantum support vector machine (qsvm) on
  near term quantum processors.'' Ph.D. dissertation, Brac University, 2019.

\bibitem{willsch2020support}
D.~Willsch, M.~Willsch, H.~De~Raedt, and K.~Michielsen, ``Support vector
  machines on the d-wave quantum annealer,'' \emph{Computer Physics
  Communications}, vol. 248, p. 107006, 2020.

\bibitem{karush1939minima}
W.~Karush, ``Minima of functions of several variables with inequalities as side
  constraints,'' \emph{M. Sc. Dissertation. Dept. of Mathematics, Univ. of
  Chicago}, 1939.

\bibitem{kuhn2014nonlinear}
H.~W. Kuhn and A.~W. Tucker, ``Nonlinear programming,'' in \emph{Traces and
  emergence of nonlinear programming}.\hskip 1em plus 0.5em minus 0.4em\relax
  Springer, 2014, pp. 247--258.

\bibitem{bottou2007support}
L.~Bottou and C.-J. Lin, ``Support vector machine solvers,'' \emph{Large scale
  kernel machines}, vol.~3, no.~1, pp. 301--320, 2007.

\bibitem{gupta2003networks}
G.~{Gupta} and M.~{Younis}, ``Load-balanced clustering of wireless sensor
  networks,'' in \emph{IEEE International Conference on Communications, 2003.
  ICC '03.}, vol.~3, 2003, pp. 1848--1852 vol.3.

\bibitem{Ghosh2005marketing}
J.~Ghosh and A.~Strehl, \emph{Clustering and Visualization of Retail Market
  Baskets}.\hskip 1em plus 0.5em minus 0.4em\relax London: Springer London,
  2005, pp. 75--102.

\bibitem{Banerjee2003documentclustering}
A.~{Banerjee} and J.~{Ghosh}, ``Competitive learning mechanisms for scalable,
  incremental and balanced clustering of streaming texts,'' in
  \emph{Proceedings of the International Joint Conference on Neural Networks,
  2003.}, vol.~4, 2003, pp. 2697--2702 vol.4.

\bibitem{UshijimaMwesigwa2017partitioning}
H.~Ushijima-Mwesigwa, C.~F.~A. Negre, and S.~M. Mniszewski, ``Graph
  partitioning using quantum annealing on the d-wave system,'' \emph{ArXiv},
  vol. abs/1705.03082, 2017.

\bibitem{Kumar2018combinatorialclustering}
V.~Kumar, G.~Bass, C.~Tomlin, and J.~Dulny, ``Quantum annealing for
  combinatorial clustering,'' \emph{Quantum Information Processing}, vol.~17,
  pp. 1--14, 2018.

\bibitem{bauckhage2018adiabatic}
C.~Bauckhage, C.~Ojeda, R.~Sifa, and S.~Wrobel, ``Adiabatic quantum computing
  for kernel k= 2 means clustering.'' in \emph{LWDA}, 2018, pp. 21--32.

\bibitem{bauckhage2019qubo}
C.~Bauckhage, N.~Piatkowski, R.~Sifa, D.~Hecker, and S.~Wrobel, ``A qubo
  formulation of the k-medoids problem.'' in \emph{LWDA}, 2019, pp. 54--63.

\bibitem{Ganganath2014samesizekclustering}
N.~{Ganganath}, C.~{Cheng}, and C.~K. {Tse}, ``Data clustering with cluster
  size constraints using a modified k-means algorithm,'' in \emph{2014
  International Conference on Cyber-Enabled Distributed Computing and Knowledge
  Discovery}, 2014, pp. 158--161.

\bibitem{Malinen2014balancedcomplexity}
M.~I. Malinen and P.~Fr{\"a}nti, ``Balanced k-means for clustering,'' in
  \emph{Structural, Syntactic, and Statistical Pattern Recognition},
  P.~Fr{\"a}nti, G.~Brown, M.~Loog, F.~Escolano, and M.~Pelillo, Eds.\hskip 1em
  plus 0.5em minus 0.4em\relax Berlin, Heidelberg: Springer Berlin Heidelberg,
  2014, pp. 32--41.

\bibitem{Inaba1994kworstcase}
\BIBentryALTinterwordspacing
M.~Inaba, N.~Katoh, and H.~Imai, ``Applications of weighted voronoi diagrams
  and randomization to variance-based k-clustering: (extended abstract),'' in
  \emph{Proceedings of the Tenth Annual Symposium on Computational Geometry},
  ser. SCG ’94.\hskip 1em plus 0.5em minus 0.4em\relax New York, NY, USA:
  Association for Computing Machinery, 1994, p. 332–339. [Online]. Available:
  \url{https://doi.org/10.1145/177424.178042}
\BIBentrySTDinterwordspacing

\end{thebibliography}

\section{Author Contributions}
P.D. contributed to the research presented in Section 4 (Linear Regression) and Section 5 (Support Vector Machine). D.A. contributed to the research presented in Section 6 (Equal-Sized k-Means Clustering). L.P. contributed in writing the introduction and conclusion of the paper. All authors reviewed and wrote the manuscript.

\section{Competing Interests}
The authors declare that there are no competing interests.

\end{document}